\documentclass[10pt,twocolumn,letterpaper]{article}

\usepackage{iccv}
\usepackage{times}
\usepackage{epsfig}
\usepackage{graphicx}
\usepackage{amsmath}
\usepackage{amssymb}
\usepackage{calligra}

\usepackage{xspace} %
\usepackage{graphicx}
\usepackage{subcaption}
\usepackage{multirow}
\usepackage{lineno}
\usepackage{booktabs}
\usepackage{bbding}        %
\usepackage[table]{xcolor} %
\usepackage{colortbl}
\usepackage{amsmath}
\usepackage{pdfrender}
\usepackage{hhline}
\usepackage[numbers]{natbib}
\definecolor{mygray}{rgb}{0.7, 0.7, 1.0}
\definecolor{mygray2}{gray}{0.9}
\definecolor{myblue}{rgb}{0.8, 0.8, 1.0}

\usepackage{xspace}
\usepackage{colortbl}

\makeatletter
\DeclareRobustCommand\onedot{\futurelet\@let@token\@onedot}
\def\@onedot{\ifx\@let@token.\else.\null\fi\xspace}

\def\eg{\emph{e.g}\onedot} 
\def\ie{\emph{i.e}\onedot} 
 
 \def\vs{\emph{vs}\onedot}

\makeatother

\usepackage[pagebackref=true,breaklinks=true,letterpaper=true,colorlinks,bookmarks=false]{hyperref}

\iccvfinalcopy %

\makeatletter
\let\@oldmaketitle\@maketitle%
\renewcommand{\@maketitle}{\@oldmaketitle%
     \centering
     \vspace{-1em}
     \includegraphics[width=\linewidth]{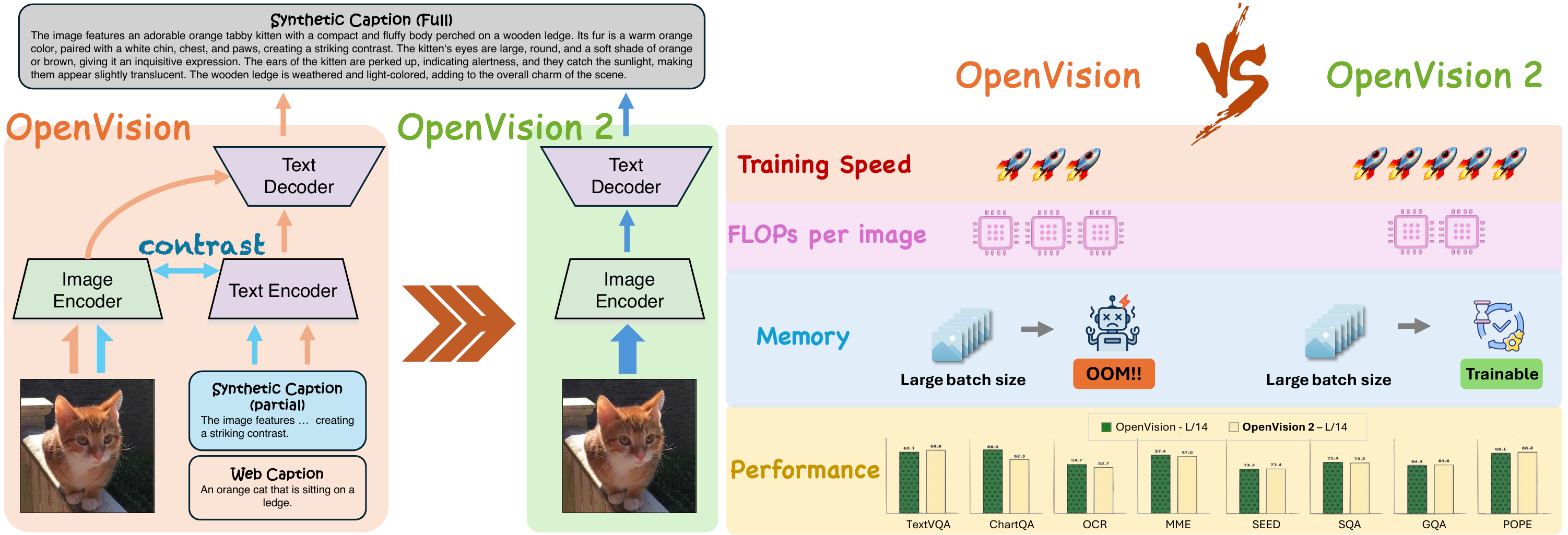}
     \vspace{-1.5em}
     \captionof{figure}{Left panel: The changes made in \textsf{OpenVision 2}. Right Panel: The benefits brought by \textsf{OpenVision 2}.}
     \label{fig:intro}
    \bigskip}%
\makeatother

\makeatletter\renewcommand\paragraph{\@startsection{paragraph}{4}{\z@}
	{.25em \@plus1ex \@minus.2ex}{-.5em}{\normalfont\normalsize\bfseries}}\makeatother
\ificcvfinal\fi

\begin{document}

\title{\textsf{OpenVision 2}\,\,\includegraphics[width=.7cm]{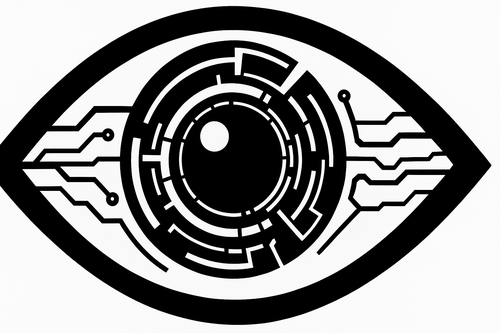}: A Family of \emph{Generative Pretrained} Visual Encoders
\\ for Multimodal Learning}

\author{%
  Yanqing Liu$^1$ \,  Xianhang Li$^1$  \, Letian Zhang$^1$ \,   Zirui Wang$^2$ \, Zeyu Zheng$^3$ \, Yuyin Zhou$^1$ \,
  Cihang Xie$^1$ \vspace{.3em}\\ 
 $^1$University of California Santa Cruz \,\, \, \,\, $^2$Apple \,\, \, \,\,$^3$University of California Berkeley \vspace{.5em}
  \\
  \small
  \hspace{3em} \includegraphics[height=1.1em]{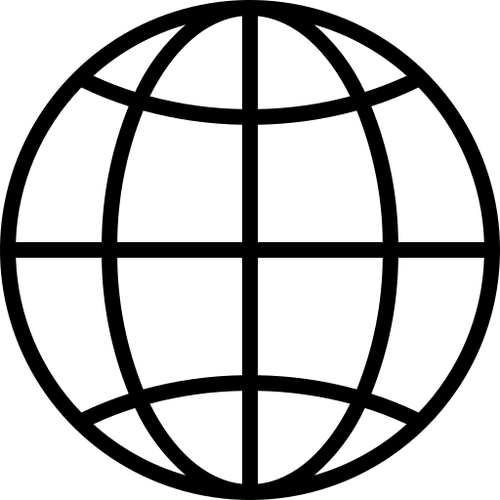} \textbf{Project Page}: \url{https://ucsc-vlaa.github.io/OpenVision2} \\
  \small
  \hspace{3em} \includegraphics[height=1.2em]{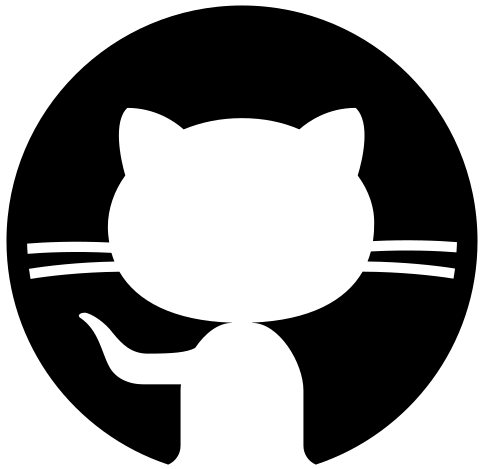} \textbf{Model Training}: \url{https://github.com/UCSC-VLAA/OpenVision} \\
  \small
  \hspace{3em} \includegraphics[height=1.2em]{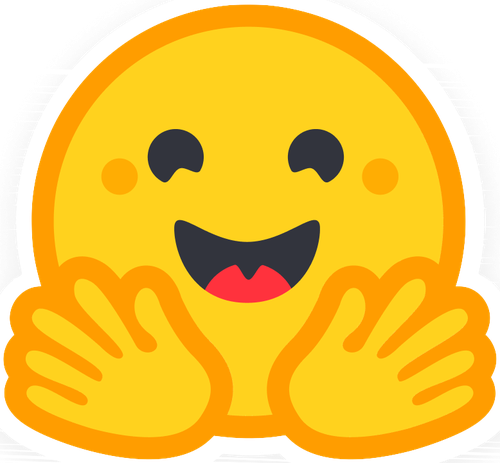} \textbf{Model Zoo}: \href{https://huggingface.co/collections/UCSC-VLAA/openvision-2-68ab5934fe21f3fc463077da}{click me}
  \hspace{3em} \includegraphics[height=1.2em]{logo_hf.png} \textbf{Recap-DataComp-1B v2}: \href{https://huggingface.co/datasets/UCSC-VLAA/Recap-DataComp-1B}{click me}
}

\maketitle
\ificcvfinal\thispagestyle{empty}\fi

\begin{abstract}
This paper provides a \textbf{simplification} on \textsf{OpenVision}'s architecture and loss design for enhancing its training efficiency. Following the prior vision-language pretraining works CapPa and AIMv2, as well as modern multimodal designs like LLaVA, our changes are straightforward: we remove the text encoder---and therefore the contrastive loss---retaining only the captioning loss as a purely generative training signal. We name this new version \textsf{OpenVision 2}. The initial results are promising: despite this simplification, \textsf{OpenVision 2} competitively matches the original model’s performance on a broad set of multimodal benchmarks while substantially cutting both training time and memory consumption. For example, with ViT-L/14, it saves the training time by $\sim1.5\times$ (\ie, from 83h to 57h), and memory usage by $\sim 1.8\times$ (\ie, from 24.5GB to 13.8GB; or equivalently, allowing the maximum batch size to grow from 2k to 8k). This superior training efficiency also allows us to scale far beyond the largest vision encoder used in \textsf{OpenVision}, reaching more than 1 billion parameters. We hold a strong belief that this lightweight, generative-only paradigm is compelling for future vision encoder development in multimodal foundation models. 

\end{abstract}

\section{Introduction}

The visual module in multimodal foundation models has long been dependent on not-fully-open solutions, like OpenAI's CLIP~\cite{radford2021clip} or Google's SigLIP~\cite{zhai2023sigmoid}.  
To mitigate this limitation, \textsf{OpenVision} \cite{li2025openvision} provides a fully open alternative: by building exclusively on the public dataset and codebase, \textsf{OpenVision} offers a family of highly competitive vision encoders,  ranging from 5.9M to 632.1M parameters, for building \textit{truly open} multimodal foundation models.

Despite its openness, the original \textsf{OpenVision} recipe is considerably heavier than the vanilla CLIP pipeline. First, its contrastive pairs are doubled: every image is paired with two captions---one web-crawled and one synthetically generated---instead of a single caption. Second, a separate generative loss (hence an extra text decoder) is added that guides the model to predict the synthetic caption given the image and the web caption. 
Although these additional overheads are largely hidden by adopting CLIPA-style training \cite{li2023clipa} (\ie, pretraining on low-resolution images followed by a short high-resolution fine-tune), trimming \textsf{OpenVision}’s overall computational footprint remains crucial for broader access, especially for researchers with limited computational resources, as well as for further scaling in data size, training epochs, and model capacity.

In this paper, building on \textsf{OpenVision}, we investigate a far simpler, and much more efficient, training recipe. Specifically, following the prior vision-language pretraining works CapPa~\cite{tschannen2024image} and AIMv2~\cite{fini2024multimodal}, and the modern multimodal designs like LLaVA~\cite{llava}, we embrace a minimalist design principle:
\begin{center}
\textit{remove the text encoder entirely.}
\end{center}
Crucially, this change also eliminates the associated training signal from contrasting image–text pairs.
As a result, the training framework now simply consists of just two parts---an image encoder and a text decoder---and learns visual representations generatively through the caption loss alone, effectively collapsing the original multi-branch pipeline into a lightweight two-module architecture and markedly reducing computational overhead.

We name this new version \textsf{OpenVision 2}.
Preliminary experiments reveal that its potential is significantly considerable. Across a suite of representative multimodal benchmarks, \textsf{OpenVision 2} effectively mirrors the performance of the original \textsf{OpenVision}. For example, under the LLaVA-Next~\cite{llavanext} framework using a ViT-L/14 backbone, the two versions post nearly identical scores on tasks such as TextVQA \cite{singh2019towards} (68.9 \vs 68.3), SQA \cite{lu2022learn} (75.2 \vs 75.4), and SEED-Bench \cite{li2023seed} (73.4 \vs 73.3), with only minor differences on tasks like OCR \cite{ocr} (537 \vs 547). More notably, efficiency emerges as the key differentiator: On ViT-L, \textsf{OpenVision 2} shortens pretraining time from 83 hours to 57 hours while cutting  computational memory from 24.5 GB to 13.8 GB per device; these gains are even more substantial with the larger SoViT-400M backbone, where training time is reduced from 241 hours to 121 hours and memory requirement from 27.4 GB to 14.5 GB per device. Importantly, this enhanced training efficiency makes it feasible to scale the vision encoder beyond 1 billion parameters---a scale that was previously less practical under the original \textsf{OpenVision} setup.
These findings altogether indicate that a caption-only generative objective can not only maintain advanced performance but also dramatically reduce computational cost and enable greater scalability.

We hope these findings prompt the community to \textit{seriously} reconsider the long-held common belief that CLIP-style contrastive learning is indispensable for building scalable, general-purpose vision encoders---a notion that prior studies such as CapPa and AIMv2 repeatedly argued for years. Specifically, in \textsf{OpenVision 2}, we demonstrate that a purely generative, caption-only objective can rival contrastive methods in multimodal performance while substantially lowering computation and memory requirements. To enable deeper and broad exploration, we release the \textit{full training suite and pretrained checkpoints} of \textsf{OpenVision 2}. We invite researchers to build on this resource and further probe the potential of \textit{generative pretraining of vision encoders} for multimodal learning.

\section{Method}

\subsection{\textsf{OpenVision}: A Review of its Technical Details}

\textsf{OpenVision} offers the research community a fully open suite to train advanced vision encoders for building multimodal foundation models.  Specifically,
compared to the vanilla CLIP setup, \textsf{OpenVision} incorporates three key changes drawn from recent literature:

\begin{itemize}
    \item \emph{Efficiency (CLIPA \cite{li2023clipa})} Training a CLIP model from scratch is prohibitively expensive. CLIPA mitigates this cost by pre-training on low-resolution images and conducting only a brief fine-tuning stage at full resolution, yielding up to a 33$\times$ speed-up. \textsf{OpenVision} adopts this two-stage curriculum to accelerate training.
    
    \item \emph{Data Quality (Recap \cite{li2024if})} Web-crawled captions are often noisy and incomplete. Recap \cite{li2024if} improves label quality by replacing them with model-generated captions. Concretely, a LLaMA-3-powered LLaVA model recaptions the entire DataComp-1B collection; this high-quality synthetic set serves as the training corpus for \textsf{OpenVision}.

    \item \emph{Optimization (CLIP\texttt{S} \cite{liu2024clips})} To better leverage synthetic captions, CLIP\texttt{S} introduces two additional objectives:  
      (i) a \emph{dual} contrastive loss that pairs each image with both web-crawled and generated captions, and  
      (ii) a caption loss that asks the model to predict the synthetic caption given the image and its web caption.  
      \textsf{OpenVision} integrates both losses to enhance training.
\end{itemize}

Combining these three ingredients enables \textsf{OpenVision} to train advanced, CLIP-style vision encoders entirely from public data with reasonable computational resources. As shown in \cite{li2025openvision}, the resulting models rival proprietary counterparts such as OpenAI’s CLIP and Google’s SigLIP in building multimodal foundation models.

\subsection{\textsf{OpenVision 2}: What are the CHANGES?}
Although the additional usage of high-quality synthetic caption enhances the overall multimodal performance, it also adds substantial computational overhead: (i) the text encoder must now process two captions per image for the dual contrastive objective, and (ii) an additional text decoder is required to autoregressively predict the synthetic caption.  Together, these two components substantially increase FLOPs and GPU memory in training.

\textsf{OpenVision 2} eliminates this computational bottleneck by discarding the text encoder---and, with it, the entire image–text contrastive loss. The training loop therefore simply collapses to two steps (left panel of Figure~\ref{fig:intro}):
\begin{enumerate}
    \item An image is processed by the vision encoder, producing a sequence of visual tokens.
    \item These tokens are passed directly to a text decoder, which predicts the paired synthetic caption.
\end{enumerate}
Viewed through this lens, \textsf{OpenVision 2} becomes a purely generative pretraining pipeline, closely mirroring the architecture used during downstream multimodal fine-tuning (\eg, LLaVA). This architectural alignment eliminates the objective mismatch between pretraining and downstream finetuning, potentially facilitating smoother knowledge transfer across stages.

On top of this, we introduce one additional efficiency tweak: during pretraining, roughly two-thirds of the visual tokens are randomly masked before they are fed to the text decoder. As confirmed in our empirical findings, the remaining one-third provides sufficient conditioning for caption generation while further reducing the text decoder’s computational load.

\paragraph{Differences from CapPa~\cite{tschannen2024image}}
\textsf{OpenVision 2} embraces the caption-only philosophy pioneered by CapPa, yet diverges and improves in the following four aspects:
\begin{enumerate}
    \item \emph{Higher-quality captions.}  CapPa is trained on short, often noisy web captions. We instead employ ReCap-DataComp-1B---a Llama-3-powered, fully re-captioned version of DataComp-1B---and \textit{further refine it with an enhanced captioning strategy\footnote{We enhance caption generation by conditioning this LLaMA-3-powered LLaVA on the original alt-text and applying weighted top-$k$ sampling, which encourages more diverse yet grounded captions. We name this new dataset Recap-DataComp-1B v2.}}. The resulting captions are longer, more grounded, and therefore better suited to generative supervision (which are empirically confirmed in Table \ref{Tab:ablation_caption}).
    \item \emph{Fusion simplification.} CapPa fuses modalities with cross-attention. We replace this with simple concatenation of visual tokens in the text decoder, following recent multimodal practice (\eg, as in LLaVA). We additionally drop a random subset of visual tokens during training, which both regularises the encoder and cuts decoding cost.
    \item \emph{Scale and evaluation scope.} Compared to CapPa, we further scale the vision encoder size up to 1.01 billion parameter trained on 12.8 B image–caption pairs. Moreover, rather than focusing on image classification and simple retrieval/QA, our vision encoders are evaluated on more advanced and complicated multimodal benchmarks like MME~\cite{fu2023mme} and ChartQA~\cite{masry-etal-2022-chartqa}.
    \item \emph{Decoding strategy.} While CapPa argues that using a mixture of extensive parallel prediction and light autoregressive prediction, we simply use its vanilla format, \ie, autoregressive prediction only, in pretraining.
\end{enumerate}

\begin{table*}[t!]
    \centering
    \caption{
    Comparison of \textsf{OpenVision 2} with existing CLIP variants under the LLaVA-1.5 framework. 
    \textsf{OpenVision 2} consistently outperforms both prior CLIP baselines and \textsf{OpenVision}, with clear gains on OCR-related tasks.
    }
    \label{Tab:llava1.5_main1}
    \vspace{-.6em}
    \resizebox{\linewidth}{!}{
    \begin{tabular}{c|c|c|c|c|c|c|c|c|c|c|c}
    \toprule
    \textbf{Method} &
    \textbf{Vision Encoder} &
    \textbf{Params} &
    \textbf{\# Res.} &
    \textbf{Text VQA} &
    \textbf{Chart QA} &
    \textbf{OCR.} &
    \textbf{MME} &
    \textbf{SEED} &
    \textbf{SQA} &
    \textbf{GQA}  &
    \textbf{POPE} \\
    \midrule
    OpenAI-CLIP~\citep{radford2021clip} & L/14 & 304M & 224 & 56.1 & 13.2 & 177 & 1443/306 & 66.0 & 73.4 & 60.8 & 85.0 \\
    LAION-2B-CLIP~\citep{openclip} & L/14 & 304M & 224 & 54.2 & 12.8 & 165 & 1434/298 & 65.5 & 76.0 & 59.0 & 84.5 \\
    DataComp-1B-CLIP~\citep{gadre2023datacomp} & L/14 & 304M & 224 & 53.0 & 12.3 & 131 & 1382/312 & 62.4 & 74.2 & 57.8 & 83.0 \\
    DFN-2B-CLIP~\citep{dfn} & L/14 & 304M & 224 & 53.2 & 12.4 & 246 & 1447/306 & 65.6 & 76.3 & 59.1 & 85.0 \\
    MetaCLIP-5B~\citep{metaclip} & L/14 & 304M & 224 & 55.6 & 12.8 & 313 & 1552/315 & 67.4 & 78.0 & 61.3 & 85.4 \\
    \rowcolor{gray!20}\textsf{OpenVision \cite{li2025openvision}} & \cellcolor{gray!20}L/14 & \cellcolor{gray!20}304M & \cellcolor{gray!20}224 & 57.7 & 13.9 & 315 & 1487/317 & 69.5 & 73.6 & 62.9 & 86.4 \\
    \rowcolor{cyan!10}\textbf{\textsf{OpenVision 2}} & \cellcolor{cyan!10}\textbf{L/14} & \cellcolor{cyan!10}\textbf{304M} & \cellcolor{cyan!10}\textbf{224} & \textbf{59.0} & \textbf{13.7} & \textbf{327} & \textbf{1460/312} & \textbf{69.3} & \textbf{76.5} & \textbf{62.6} & \textbf{87.1} \\
    \midrule
    OpenAI-CLIP~\citep{radford2021clip} & L/14 & 304M & 336 & 59.1 & 13.8 & 201 & 1475/288 & 67.5 & 73.1 & 61.1 & 85.7 \\
    \rowcolor{gray!20}\textsf{OpenVision \cite{li2025openvision}} & \cellcolor{gray!20}L/14 & \cellcolor{gray!20}304M & \cellcolor{gray!20}336 & 61.2 & 15.7 & 339 & 1525/315 & 70.5 & 75.1 & 63.7 & 87.2 \\
    \rowcolor{cyan!10}\textbf{\textsf{OpenVision 2}} & \cellcolor{cyan!10}\textbf{L/14} & \cellcolor{cyan!10}\textbf{304M} & \cellcolor{cyan!10}\textbf{336} & \textbf{63.0} & \textbf{14.5} & \textbf{357} & \textbf{1486/321} & \textbf{70.1} & \textbf{77.5} & \textbf{63.0} & \textbf{87.7} \\
    \midrule
    SigLIP~\citep{zhai2023sigmoid} & SoViT-400M/14 & 400M & 384 & 62.6 & 14.5 & 338 & 1481/347 & 69.4 & 76.7 & 63.3 & 87.0 \\
    \rowcolor{gray!20}\textsf{OpenVision \cite{li2025openvision}} & \cellcolor{gray!20}SoViT-400M/14 & \cellcolor{gray!20}400M & \cellcolor{gray!20}384 & 62.4 & 16.1 & 357 & 1493/320 & 70.4 & 72.4 & 63.8 & 88.0 \\
    \rowcolor{cyan!10}\textbf{\textsf{OpenVision 2}} & \cellcolor{cyan!10}\textbf{SoViT-400M/14} & \cellcolor{cyan!10}\textbf{400M} & \cellcolor{cyan!10}\textbf{384} & \textbf{64.3} & \textbf{15.0} & \textbf{387} & \textbf{1472/310} & \textbf{70.7} & \textbf{74.9} & \textbf{63.5} & \textbf{87.5} \\
    \midrule
    \rowcolor{cyan!10}\textbf{\textsf{OpenVision 2}} & \cellcolor{cyan!10}\textbf{H/14} & \cellcolor{cyan!10}\textbf{632M} & \cellcolor{cyan!10}\textbf{224} & \textbf{60.2} & \textbf{13.5} & \textbf{340} & \textbf{1470/305} & \textbf{69.3} & \textbf{75.4} & \textbf{62.5} & \textbf{87.2} \\
    \rowcolor{cyan!10}\textbf{\textsf{OpenVision 2}} & \cellcolor{cyan!10}\textbf{H/14} & \cellcolor{cyan!10}\textbf{632M} & \cellcolor{cyan!10}\textbf{336} & \textbf{63.4} & \textbf{16.3} & \textbf{391} & \textbf{1470/311} & \textbf{70.6} & \textbf{76.4} & \textbf{63.1} & \textbf{88.4} \\
    \rowcolor{cyan!10}\textbf{\textsf{OpenVision 2}} & \cellcolor{cyan!10}\textbf{H/14} & \cellcolor{cyan!10}\textbf{632M} & \cellcolor{cyan!10}\textbf{448} & \textbf{65.6} & \textbf{18.1} & \textbf{416} & \textbf{1499/331} & \textbf{70.6} & \textbf{75.6} & \textbf{63.1} & \textbf{88.7} \\
    \midrule
    \rowcolor{cyan!10}\textbf{\textsf{OpenVision 2}} & \cellcolor{cyan!10}\textbf{g/14} & \cellcolor{cyan!10}\textbf{1.01B} & \cellcolor{cyan!10}\textbf{224} & \textbf{60.2} & \textbf{13.7} & \textbf{338} & \textbf{1469/290} & \textbf{69.3} & \textbf{75.0} & \textbf{62.6} & \textbf{86.9} \\
    \bottomrule
    \end{tabular}}
    \vspace{-.5em}
\end{table*}

\begin{table*}[t!]
    \centering
    \caption{
    Comparison of \textsf{OpenVision 2} with existing CLIP variants under the Open-LLaVA-Next framework. 
    \textsf{OpenVision 2} achieves competitive or better results across model scales, surpassing prior CLIP baselines and \textsf{OpenVision}.
    }
    \label{Tab:llava-next_main1}
    \vspace{-.6em}
    \resizebox{\linewidth}{!}{
    \begin{tabular}{c|c|c|c|c|c|c|c|c|c|c|c}
    \toprule
    \textbf{Method} & \textbf{Vision Encoder} & \textbf{Params} & \textbf{\# Res.} & \textbf{Text VQA} & \textbf{Chart QA} & \textbf{OCR.} & \textbf{MME} & \textbf{SEED} & \textbf{SQA} & \textbf{GQA} & \textbf{POPE} \\
    \midrule
    OpenAI-CLIP~\citep{radford2021clip} & L/14 & 304M & 224 & 62.8 & 60.7 & 459 & 1600/334 & 70.6 & 75.0 & 62.8 & 86.9 \\
    LAION-2B-CLIP~\citep{openclip} & L/14 & 304M & 224 & 59.4 & 50.8 & 396 & 1533/323 & 70.0 & 72.9 & 62.7 & 86.4 \\
    DataComp-1B-CLIP~\citep{gadre2023datacomp} & L/14 & 304M & 224 & 58.1 & 48.5 & 373 & 1524/348 & 70.2 & 75.6 & 62.3 & 86.2 \\
    DFN-2B-CLIP~\citep{dfn} & L/14 & 304M & 224 & 57.0 & 42.7 & 303 & 1486/328 & 68.3 & 70.6 & 61.7 & 86.0 \\
    MetaCLIP-5B~\citep{metaclip} & L/14 & 304M & 224 & 63.0 & 62.9 & 493 & 1590/335 & 72.3 & 77.1 & 64.0 & 86.8 \\
    \rowcolor{gray!20}\textsf{OpenVision} & \cellcolor{gray!20}L/14 & \cellcolor{gray!20}304M & \cellcolor{gray!20}224 & \cellcolor{gray!20}65.7 & \cellcolor{gray!20}61.5 & \cellcolor{gray!20}503 & \cellcolor{gray!20}1567/332 & \cellcolor{gray!20}73.1 & \cellcolor{gray!20}73.1 & \cellcolor{gray!20}64.7 & \cellcolor{gray!20}87.8 \\
    \rowcolor{cyan!10}\textbf{\textsf{OpenVision 2}} & \cellcolor{cyan!10}\textbf{L/14} & \cellcolor{cyan!10}\textbf{304M} & \cellcolor{cyan!10}\textbf{224} & \textbf{66.1} & \textbf{60.4} & \textbf{501} & \textbf{1577/297} & \textbf{73.1} & \textbf{68.4} & \textbf{64.6} & \textbf{87.6} \\
    \midrule
    OpenAI-CLIP~\citep{radford2021clip} & L/14 & 304M & 336 & 69.4 & 70.0 & 535 & 1591/351 & 73.3 & 76.9 & 64.5 & 87.6 \\
    \rowcolor{gray!20}\textsf{OpenVision} & \cellcolor{gray!20}L/14 & \cellcolor{gray!20}304M & \cellcolor{gray!20}336 & \cellcolor{gray!20}68.3 & \cellcolor{gray!20}68.0 & \cellcolor{gray!20}547 & \cellcolor{gray!20}1520/310 & \cellcolor{gray!20}73.3 & \cellcolor{gray!20}75.4 & \cellcolor{gray!20}64.4 & \cellcolor{gray!20}88.1 \\
    \rowcolor{cyan!10}\textbf{\textsf{OpenVision 2}} & \cellcolor{cyan!10}\textbf{L/14} & \cellcolor{cyan!10}\textbf{304M} & \cellcolor{cyan!10}\textbf{336} & \textbf{68.9} & \textbf{62.3} & \textbf{537} & \textbf{1585/278} & \textbf{73.4} & \textbf{75.2} & \textbf{64.6} & \textbf{88.4} \\
    \midrule
    SigLIP~\citep{zhai2023sigmoid} & SoViT-400M/14 & 400M & 384 & 68.2 & 61.3 & 494 & 1539/325 & 72.9 & 74.7 & 62.9 & 86.8 \\
    \rowcolor{gray!20}\textsf{OpenVision} & \cellcolor{gray!20}SoViT-400M/14 & \cellcolor{gray!20}400M & \cellcolor{gray!20}384 & \cellcolor{gray!20}67.4 & \cellcolor{gray!20}63.1 & \cellcolor{gray!20}540 & \cellcolor{gray!20}1500/353 & \cellcolor{gray!20}72.2 & \cellcolor{gray!20}73.5 & \cellcolor{gray!20}63.4 & \cellcolor{gray!20}87.8 \\
    \rowcolor{cyan!10}\textbf{\textsf{OpenVision 2}} & \cellcolor{cyan!10}\textbf{SoViT-400M/14} & \cellcolor{cyan!10}\textbf{400M} & \cellcolor{cyan!10}\textbf{384} & \textbf{69.0} & \textbf{63.4} & \textbf{549} & \textbf{1521/319} & \textbf{72.2} & \textbf{72.7} & \textbf{63.1} & \textbf{87.7} \\
    \midrule
    \rowcolor{cyan!10}\textbf{\textsf{OpenVision 2}} & \cellcolor{cyan!10}\textbf{H/14} & \cellcolor{cyan!10}\textbf{632M} & \cellcolor{cyan!10}\textbf{224} & \textbf{66.4} & \textbf{60.2} & \textbf{514} & \textbf{1597/314} & \textbf{73.3} & \textbf{76.2} & \textbf{64.7} & \textbf{88.4} \\
    \rowcolor{cyan!10}\textbf{\textsf{OpenVision 2}} & \cellcolor{cyan!10}\textbf{H/14} & \cellcolor{cyan!10}\textbf{632M} & \cellcolor{cyan!10}\textbf{336} & \textbf{69.9} & \textbf{64.8} & \textbf{573} & \textbf{1572/337} & \textbf{73.8} & \textbf{74.5} & \textbf{64.4} & \textbf{87.8} \\
    \rowcolor{cyan!10}\textbf{\textsf{OpenVision 2}} & \cellcolor{cyan!10}\textbf{H/14} & \cellcolor{cyan!10}\textbf{632M} & \cellcolor{cyan!10}\textbf{448} & \textbf{71.9} & \textbf{64.9} & \textbf{590} & \textbf{1542/324} & \textbf{74.1} & \textbf{75.6} & \textbf{64.4} & \textbf{88.8} \\
    \midrule
    \rowcolor{cyan!10}\textbf{\textsf{OpenVision 2}} & \cellcolor{cyan!10}\textbf{g/14} & \cellcolor{cyan!10}\textbf{1.01B} & \cellcolor{cyan!10}\textbf{224} & \textbf{67.3} & \textbf{62.4} & \textbf{514} & \textbf{1558/323} & \textbf{73.4} & \textbf{74.4} & \textbf{64.7} & \textbf{88.0} \\
    \bottomrule
    \end{tabular}}
    \vspace{-.7em}
\end{table*}

\paragraph{Differences from AIMv2~\cite{fini2024multimodal}} 
The overall design of our \textsf{OpenVision 2} is more closed to the more recent AIMv2, but still differs in the following ways:
\begin{enumerate}
    \item \textit{Training signal.} AIMv2 supervises the vision encoder with a multimodal autoregressive decoder that \emph{simultaneously} (i) reconstructs image patches through pixel-level regression and (ii) generates text tokens, blending image-level and text-level objectives. \textsf{OpenVision 2}, in contrast, follows CapPa’s caption-only philosophy: textual generation is the only learning signal, and no image-reconstruction loss is introduced.

    \item \textit{Token-masking scheme.} Compared to AIMv2, \textsf{OpenVision 2} randomly masks roughly two-thirds of the visual tokens before passing them to the text decoder. As confirmed in our empirical results, this design enhances both the training efficiency and the multimodal performance.

    \item \textit{Data composition.} AIMv2 is trained on a mix of human and synthetic captions ($\sim 67\%$ real, $33\%$ synthetic). Our corpus is entirely synthetic and produced with the ReCap-DataComp-1B pipeline, yielding richer, more consistent descriptions that better align with a purely generative objective.

    \item \textit{Vision encoder architecture.} AIMv2 adopts a prefix-ViT~\cite{el2024scalable}, where the attention mask allows prefix tokens to attend bidirectionally while the remaining tokens are modeled autoregressively. In contrast, \textsf{OpenVision 2} simply uses a vanilla ViT backbone without such modifications, keeping the encoder simple and efficient.

\end{enumerate}

\section{Results}
\subsection{Multimodal Benchmark Performance}
\label{sec:results_multimodal}

Following \textsf{OpenVision}, we evaluate the effectiveness of  \textsf{OpenVision 2} on a range of multimodal downstream tasks, under both the LLaVA-1.5~\cite{liu2023improved} and Open-LLaVA-Next~\cite{chen2024open} frameworks. Specifically, we report results on commonly used multimodal benchmarks, MME~\cite{fu2023mme}, GQA~\cite{hudson2018gqa}, ChartQA~\cite{masry-etal-2022-chartqa}, POPE~\cite{Li-hallucination-2023}, TextVQA~\cite{singh2019towards}, OCR~\cite{ocr}, SEED~\cite{li2023seed}, and SQA~\cite{lu2022learn}. 
The results are summarized in Table~\ref{Tab:llava1.5_main1} and Table~\ref{Tab:llava-next_main1}.

\paragraph{LLaVA-1.5 results.}  
As shown in Table~\ref{Tab:llava1.5_main1}, \textsf{OpenVision 2} achieves performance comparable to or better than \textsf{OpenVision} models, while being significantly more efficient in terms of training time and memory (see Sec.~\ref{sec:efficiency}). 
For instance, at the ViT-L/14 resolution-224 setting, \textsf{OpenVision 2} matches or slightly exceeds \textsf{OpenVision} (59.0 \vs 57.7 on TextVQA, 13.7 \vs 13.9 on ChartQA, and 327 \vs 315 on OCR-Bench), despite reducing the training cost by $\sim$1.5$\times$. 
The same trend holds for larger scales (\eg, SoViT-400M/14, H/14), where \textsf{OpenVision 2} preserves the strong performance of \textsf{OpenVision} while offering markedly better efficiency.  
This demonstrates that our efficiency-oriented design does not compromise accuracy, even under more challenging benchmarks such as OCR-related tasks.  

\paragraph{Open-LLaVA-Next results.}  
A similar pattern is observed under the Open-LLaVA-Next framework (Table~\ref{Tab:llava-next_main1}). 
\textsf{OpenVision 2} consistently delivers results on par with, or better than, the original \textsf{OpenVision}, while retaining its large efficiency advantage. 
For example, at the ViT-L/14 resolution-336 setting, \textsf{OpenVision 2} achieves 68.9 on TextVQA, 537 on OCR-Bench, and 1585 on MME-Perception, closely matching or slightly improving upon \textsf{OpenVision} (68.3, 547, and 1520, respectively). 
When scaling to SoViT-400M/14 and H/14, \textsf{OpenVision 2} further strengthens this trend, matching the strong baselines set by \textsf{OpenVision} and in some cases surpassing them (\eg, +19 on OCR-Bench over OpenVision with SoViT-400M/14, and new best results on TextVQA, OCR-Bench, and POPE with H/14-448).  
These results confirm that efficiency improvements in \textsf{OpenVision 2} come without sacrificing—and sometimes even enhancing—downstream performance.

\paragraph{Overall trends.}  
These results demonstrate three key takeaways:  
(i) \textsf{OpenVision 2} generalizes well across both multimodal frameworks, showing consistent improvements over previous CLIP models.  
(ii) The advantage is particularly strong on OCR-intensive benchmarks, validating the effectiveness of our synthetic captioning and token masking strategies for enhancing fine-grained text recognition.  
(iii) The improvements scale smoothly with both model size and input resolution, confirming that \textsf{OpenVision 2} maintains efficiency and robustness under large-scale settings.  
(iv) Compared to the first version of \textsf{OpenVision}, our new design achieves these gains while substantially reducing training time and memory footprint (see Sec.~\ref{sec:efficiency}).  

\subsection{Training Efficiency and Scalability}
\label{sec:efficiency}

A key advantage of \textsf{OpenVision 2} lies in its superior training efficiency and scalability. 
All experiments are conducted on Google Cloud TPUs, with training time measured on v4-512 pods and memory usage on v4-64 pods. 
We report wall-clock training time, computational cost, and memory footprint across different model sizes in Table~\ref{tab:efficiency_time}, Table~\ref{tab:efficiency_memory_scaling}, and Table~\ref{Tab:efficiency}.  

\begin{table*}[t]
\centering
\caption{Training efficiency of \textsf{OpenVision} variants on TPU v4-512. \textbf{\textsf{OpenVision 2} achieves faster training and lower computational cost across model sizes.}}
\label{tab:efficiency_time}
\vspace{-0.5em}
\resizebox{0.65\linewidth}{!}{
\begin{tabular}{l|c|c|c|c}
\toprule
\textbf{Model} & \textbf{Backbone} & \textbf{Resolution} & \textbf{v4-512 Hours} & \textbf{FLOPs / Image} \\
\midrule
\textsf{OpenVision \cite{li2025openvision}} & L/14 & 224 & 83 & 271.75 \\
\textbf{\textsf{OpenVision 2}} & \textbf{L/14} & \textbf{224} & \textbf{57} & \textbf{208.90} \\
\midrule
\textsf{OpenVision \cite{li2025openvision}} & SoViT-400M/14 & 384 & 241 & 1636.75 \\
\textbf{\textsf{OpenVision 2}} & \textbf{SoViT-400M/14} & \textbf{384} & \textbf{121} & \textbf{1017.74} \\
\bottomrule
\end{tabular}}
\vspace{-0.5em}
\end{table*}

\begin{table*}[t]
\centering
\caption{
Peak memory usage (GB, measured per TPU chip) of \textsf{OpenVision} variants on TPU v4-64 under increasing batch sizes. 
\textbf{\textsf{OpenVision 2} supports larger batches with significantly lower memory footprint.}
}
\label{tab:efficiency_memory_scaling}
\vspace{-0.5em}
\resizebox{0.65\linewidth}{!}{
\begin{tabular}{l|c|c|c}
\toprule
\textbf{Model} & \textbf{Resolution} & \textbf{Batch Size} & \textbf{Peak Memory (GB)} \\
\midrule
\multirow{2}{*}{\textsf{OpenVision \cite{li2025openvision} (L/14)}} 
& 224 & 2k & 24.5 \\
& 224 & 4k & \textcolor{gray}{OOM} \\
\cmidrule{1-4}
\multirow{3}{*}{\textbf{\textsf{OpenVision 2 (L/14)}}} 
& \textbf{224} & \textbf{2k} & \textbf{13.8} \\
& \textbf{224} & \textbf{4k} & \textbf{22.1} \\
& \textbf{224} & \textbf{8k} & \textbf{28.4} \\
\midrule
\multirow{2}{*}{\textsf{OpenVision \cite{li2025openvision} (SoViT-400M/14)}} 
& 384 & 512 & 27.4 \\
& 384 & 1k & \textcolor{gray}{OOM} \\
\cmidrule{1-4}
\multirow{2}{*}{\textbf{\textsf{OpenVision 2 (SoViT-400M/14)}}} 
& \textbf{384} & \textbf{512} & \textbf{14.5} \\
& \textbf{384} & \textbf{1k} & \textbf{28.8} \\
\bottomrule
\end{tabular}}
\vspace{-0.5em}
\end{table*}

\begin{table*}[t!]
    \centering
    \caption{Training efficiency comparison on ViT-L/14 @224, measured on TPU v4-64.
    We report wall-clock training time (hours) and indicate whether CLIPA optimization or masked-token strategy is applied.}
    \label{Tab:efficiency}
    \vspace{-.6em}
    \resizebox{0.65\linewidth}{!}{
    \begin{tabular}{c|c|c|c}
    \toprule
    \textbf{Method} & \textbf{CLIPA Opt.} & \textbf{Mask Token Opt.} & \textbf{Training Time (h)} \\
    \midrule
    CapPa~\cite{tschannen2024image} (baseline) & -- & -- & 217 \\
    \textsf{OpenVision 2 (w/ Mask only)} & -- & \checkmark & 190 \\
    \textsf{OpenVision 2 (w/ CLIPA only)} & \checkmark & -- & 67 \\
    \textsf{OpenVision 2 (w/ both)} & \checkmark & \checkmark & 55 \\
    \bottomrule
    \end{tabular}}
    \vspace{-.5em}
\end{table*}

\paragraph{Training time and FLOPs.}  
Compared to the first version of \textsf{OpenVision}, our new design significantly reduces the training cost. 
For example, with ViT-L/14 at resolution 224, training time is reduced from \textbf{83h} to \textbf{57h} (\textbf{$\sim1.5\times$} faster), while the per-image FLOPs drop from 271.8 to 208.9 (\textbf{$\sim1.3\times$} lower). 
Similarly, with SoViT-400M/14 at resolution 384, training time drops from \textbf{241h} to \textbf{121h} ($\sim2\times$ faster), with per-image FLOPs reduced from 1636.8 to 1017.7.
Importantly, these efficiency gains are achieved while maintaining comparable or even stronger performance on multimodal benchmarks (see Sec.~\ref{sec:results_multimodal}).  

\paragraph{Memory Analysis.}  
\textsf{OpenVision 2} also demonstrates substantial memory savings, enabling much larger batch sizes. 
As shown in Table~\ref{tab:efficiency_memory_scaling}, at the ViT-L/14 resolution-224 setting, the peak memory usage per TPU chip drops from 24.5GB to 13.8GB (\textbf{$\sim1.8\times$} lower) at batch size 2k. 
Equivalently, the maximum batch size increases from 2k to 8k, while still remaining within the 32GB memory limit of TPU v4 cores. 
A similar trend is observed for SoViT-400M/14 at resolution 384, where \textsf{OpenVision 2} supports batch size 1k, whereas the previous version fails with out-of-memory (OOM).  

\paragraph{Effect of CLIPA and token masking.}  
To disentangle the contribution of each optimization, we further compare CapPa~\cite{tschannen2024image} and our variants in Table~\ref{Tab:efficiency}. 
Both CLIPA optimization and token masking contribute to efficiency improvements individually, while their combination yields the best results. 
For instance, on ViT-L/14 at resolution 224, the training time reduces from 217h (CapPa baseline) to 190h with masked tokens only, to 67h with CLIPA optimization only, and further to 55h when both strategies are combined. 
This demonstrates that our design not only improves efficiency, but also scales synergistically when the two optimizations are applied together.  

\paragraph{Summary.}  
In summary, \textsf{OpenVision 2} achieves significant reductions in training time, FLOPs, and memory footprint, enabling efficient scaling to high-resolution inputs and larger batch sizes. 
These efficiency gains further make it feasible to push vision encoders to the billion-parameter regime. 
Indeed, we successfully train a \textsf{OpenVision 2-g/14} model with 1B parameters (see Table~\ref{Tab:llava1.5_main1} and Table~\ref{Tab:llava-next_main1}), which sets strong new baselines while maintaining cost-effectiveness.

\begin{table*}[t!]
    \centering
    \caption{Ablation study of \textsf{OpenVision 2} trained with different caption types. 
Alt-text: raw web alt-text; 
ReCap-DataComp-1B: synthetic captions generated by MLLM without conditioning on alt-text; 
ReCap-DataComp-1B v2: synthetic captions generated by MLLM with both image and alt-text as input.}
    \label{Tab:ablation_caption}
    \vspace{-.6em}
    \resizebox{0.8\linewidth}{!}{
    \begin{tabular}{c|c|c|c|c|c|c|c|c}
    \toprule
    \textbf{Caption Type} & \textbf{Text VQA} & \textbf{Chart QA} & \textbf{OCR.} & \textbf{MME} & \textbf{SEED} & \textbf{SQA} & \textbf{GQA} & \textbf{POPE} \\
    \midrule
    Alt-text & 51.8 & 12.3 & 238 & 1306/293 & 58.6 & 75.3 & 55.4 & 82.2 \\
    ReCap-DataComp-1B & 56.9 & 12.9 & 291 & 1426/293 & 67.9 & 74.5 & 61.9 & 86.5 \\
    ReCap-DataComp-1B v2 & 56.5 & 13.1 & 303 & 1451/310 & 67.8 & 74.7 & 61.2 & 86.6 \\
    \bottomrule
    \end{tabular}}
    \vspace{-.5em}
\end{table*}

\begin{table*}[t!]
    \centering
    \caption{Ablation study of \textsf{OpenVision 2} with different image token \textbf{keep ratios}.
    A higher keep ratio retains more vision tokens as captioning conditions, while a lower keep ratio masks more tokens.}
    \label{Tab:ablation_tokenkeep}
    \vspace{-.6em}
    \resizebox{0.7\linewidth}{!}{
    \begin{tabular}{c|c|c|c|c|c|c|c|c}
    \toprule
    \textbf{Keep Ratio} & \textbf{Text VQA} & \textbf{Chart QA} & \textbf{OCR.} & \textbf{MME} & \textbf{SEED} & \textbf{SQA} & \textbf{GQA} & \textbf{POPE} \\
    \midrule
    100\%   & 53.8 & 12.2 & 254 & 1409/350 & 65.9 & 73.9 & 60.3 & 84.7 \\
     90\% & 56.3 & 12.4 & 266 & 1461/335 & 67.6 & 74.8 & 61.1 & 85.4 \\
     75\%   & 55.8 & 13.1 & 293 & 1438/283 & 68.6 & 73.9 & 61.7 & 86.3 \\
     50\%   & 55.4 & 12.8 & 299 & 1429/313 & 68.5 & 73.8 & 61.6 & 86.5 \\
     35\%   & 56.9 & 12.9 & 291 & 1426/293 & 67.9 & 74.5 & 61.9 & 86.5 \\
     25\%   & 56.7 & 12.5 & 283 & 1430/297 & 67.8 & 76.3 & 61.4 & 86.3 \\
     10\%   & 55.6 & 13.0 & 276 & 1412/301 & 66.1 & 75.0 & 61.2 & 85.4 \\
    \bottomrule
    \end{tabular}}
    \vspace{-.5em}
\end{table*}

\subsection{Ablation Studies}
\label{sec:ablation}

We conduct ablation studies to better understand the design choices of \textsf{OpenVision 2}. 
In particular, we focus on the effect of caption supervision and the masking ratio of image tokens. 
Results are summarized in Table~\ref{Tab:ablation_caption} and Table~\ref{Tab:ablation_tokenkeep}.

\paragraph{Effect of caption type.}  
We first analyze the impact of different caption sources used for training. 
As shown in Table~\ref{Tab:ablation_caption}, models trained on raw alt-text perform the worst, reflecting the noisiness and inconsistency of web annotations. 
Replacing raw alt-text with synthetic captions generated by MLLMs leads to significant improvements. 
Both ReCap-DataComp-1B and its variant ReCap-DataComp-1B v2 substantially outperform raw alt-text, with gains of \textbf{+5.1} on TextVQA and \textbf{+53} on OCR-Bench. 
While ReCap-DataComp-1B achieves slightly higher scores on certain benchmarks, we adopt ReCap-DataComp-1B v2 as our default setting. 
This choice is motivated by two factors: (i) v2 shows stronger performance on OCR-related tasks, which are critical for multimodal benchmarks, and (ii) by conditioning on raw alt-text during caption generation, v2 provides additional diversity and latent knowledge that benefit tasks requiring broader reasoning (\eg, ScienceQA~\cite{lu2022learn}). 
These results highlight the effectiveness of large-scale synthetic captioning as a reliable supervisory signal for vision-language pretraining.

\paragraph{Effect of image token keep ratio.}  
We further investigate the impact of masking strategy by varying the proportion of vision tokens retained as captioning conditions. 
Table~\ref{Tab:ablation_tokenkeep} shows that keeping all tokens (100\%) does not yield the best results; instead, moderate masking ratios lead to stronger performance. 
In particular, retaining only 25--35\% of tokens strikes the best balance, improving OCR-Bench and TextVQA compared to both extremes (100\% or 10\%). 
This confirms token masking not only reduces training overhead, but also enhances local semantic representations by forcing the model to rely on fewer, more informative visual tokens.

\section{Related Work}
\paragraph{Vision-Language Pretraining.}
Since CLIP~\cite{radford2021clip} and ALIGN~\cite{jia2021align}, contrastive learning with noisy web alt-text has become the mainstream paradigm for vision–language pretraining.
This trend was further reinforced by open datasets such as LAION-400M~\cite{schuhmann2021laion}.
Follow-up works explored more efficient architectures and alignment strategies, including ALBEF~\cite{li2021align} with momentum distillation and cross-attention, ViLT~\cite{kim2021vilt} that directly embeds image patches into a Transformer, and Pixel-BERT~\cite{huang2020pixel} with pixel-level alignment. 
More recent efforts improved scalability and robustness through data filtering and training strategies, such as EVA-CLIP~\cite{EVA-CLIP}, DataComp~\cite{gadre2023datacomp} and DFN~\cite{fang2023data}. 
Recent data-centric approaches further improve supervision by refining or synthesizing captions, 
such as LaCLIP~\cite{laclip}, VeCLIP~\cite{lai2024veclip}, DreamLIP~\cite{dreamlip}, and Liu et al.~\cite{liu2023mllms}.  
Meanwhile, BLIP~\cite{blip} and BLIP-2~\cite{blip2} incorporated caption filtering and LLM connectors (Q-Former), while PaLI/PaLI-X~\cite{chen2022pali,chen2023pali} and Kosmos-2~\cite{peng2023kosmos} emphasized multilingual scaling and grounding. 
OpenVision~\cite{li2025openvision} introduced the first fully open and cost-effective family of vision encoders, showing strong results under both LLaVA-1.5 and Open-LLaVA-Next. 
These advances established contrastive pretraining as a strong paradigm, but left open questions on generative supervision and efficiency at scale. 

\paragraph{Generative Pretraining.}
Generative modeling has become a central paradigm for multimodal learning, inspired by autoregressive language models such as GPT~\cite{radford2018improving,radford2019language,brown2020language}. 
On the vision side, iGPT~\cite{chen2020generative} treated pixels as tokens, while SimVLM~\cite{simvlm} introduced a prefix-LM objective for weakly supervised pretraining. 
More recent works integrate caption generation as supervision: CoCa~\cite{yu2022coca} combined contrastive and generative losses, Flamingo~\cite{alayrac2022flamingo} connected frozen image encoders with LLMs via cross-attention, and CapPa~\cite{tschannen2024image} advocated caption-only pretraining. 
AIMv2~\cite{fini2024multimodal} further employed multimodal autoregression with Prefix-ViT, while large-scale models such as Emu~\cite{sun2023emu}, Chameleon~\cite{team2024chameleon}, Unified-IO 2~\cite{lu2024unified}, VILA-U~\cite{wu2024vila}, and related extensions~\cite{diao2025temporal,diao2025learning} unified generation across diverse modalities. 
These generative approaches demonstrate stronger synergy with language modeling, but often increase training cost. 
Our work follows this line, while simplifying the architecture and emphasizing efficiency.  
\paragraph{Image Captioning.}
In addition to contrastive learning, captioning offers a complementary path for visual–language representation learning, serving both as a benchmark task and a source of pretraining signal.
Early works, from encoder–decoder models~\cite{vinyals2015show,xu2015show} to region-level and weakly supervised approaches~\cite{karpathy2015deep,joulin2016learning,li2017learning,sariyildiz2020learning,desai2021virtex}, established the foundations of captioning.
More recent models leverage web-scale caption generation within multimodal pretraining, such as GIT~\cite{wang2022git}, BLIP~\cite{blip}, and Show-o~\cite{xie2024show}. 
Recent studies show that captioners themselves can serve as scalable vision learners~\cite{tschannen2024image}, motivating caption-only pretraining as an alternative to contrastive objectives.
Our work continues this trajectory by leveraging synthetic captions at scale, demonstrating their effectiveness as the sole supervisory signal.

\section{Conclusion}
This work introduces \textsf{OpenVision 2}, which provides a significant simplification to \textsf{OpenVision} by training solely with the caption loss. Corroborated with prior works, we, once again, challenge the prevailing belief that CLIP-style contrastive learning is indispensable for scalable, general-purpose vision encoders; instead, we demonstrate that caption-only, generative pretraining is not only viable but often a preferable alternative.
To catalyze further work, we release the full OpenVision 2 training code, pretrained checkpoints, and the ReCap-DataComp-1B v2 corpus. We invite the community to build on this resource and to explore the broader design space of generative pretraining paradigm to build vision encoder for multimodal foundation models.

\section*{Acknowledgment}
We would like to thank TPU Research Cloud (TRC) program and Google Cloud Research Credits program for supporting our computing needs.

{\small
\bibliographystyle{ieee_fullname}
\bibliography{egbib}
}

\end{document}